\documentclass{article}

\usepackage[preprint,nonatbib]{neurips_2022}

\usepackage[utf8]{inputenc} 
\usepackage[T1]{fontenc}    
\usepackage{hyperref}       
\usepackage{url}            
\usepackage{booktabs}       
\usepackage{amsfonts}       
\usepackage{nicefrac}       
\usepackage{microtype}      
\usepackage{xcolor}         
\usepackage{graphicx}
\usepackage{enumerate}
\usepackage{wrapfig}
\usepackage{caption}
\usepackage{amsmath}
\usepackage{authblk}
\title{A Unified and Biologically-Plausible Relational Graph Representation of Vision Transformers}

\author[1]{Yuzhong Chen}
\author[2]{Yu Du}
\author[1]{Zhenxiang Xiao}
\author[3]{Lin Zhao}
\author[4]{Lu Zhang}
\author[3]{David Weizhong Liu}
\author[4]{Dajiang Zhu}
\author[2]{Tuo Zhang}
\author[2]{Xintao Hu}
\author[3]{Tianming Liu}
\author[1]{Jiang Xi \thanks{Corresponding author: xijiang@uestc.edu.cn}}
\affil[1]{University of Electronic Science and Technology of China}
\affil[2]{Northwestern Polytechnical University}
\affil[3]{University of Georgia}
\affil[4]{University of Texas at Arlington}

\begin{document}

\maketitle

\begin{abstract}
Vision transformer (ViT) and its variants have achieved remarkable successes in various visual tasks. The key characteristic of these ViT models is to adopt different aggregation strategies of spatial patch information within the artificial neural networks (ANNs). However, there is still a key lack of unified representation of different ViT architectures for systematic understanding and assessment of model representation performance. Moreover, how those well-performing ViT ANNs are similar to real biological neural networks (BNNs) is largely unexplored. To answer these fundamental questions, we, for the first time, propose a unified and biologically-plausible \emph{relational graph} representation of ViT models. Specifically, the proposed \emph{relational graph} representation consists of two key sub-graphs: \emph{aggregation graph} and \emph{affine graph}. The former one considers ViT tokens as nodes and describes their spatial interaction, while the latter one regards network channels as nodes and reflects the information communication between channels. Using this unified relational graph representation, we found that: a) a sweet spot of the \emph{aggregation graph} leads to ViTs with significantly improved predictive performance; b) the graph measures of clustering coefficient and average path length are two effective indicators of model prediction performance, especially when applying on the datasets with small samples; c) our findings are consistent across various ViT architectures and multiple datasets; d) the proposed relational graph representation of ViT has high similarity with real BNNs derived from brain science data. Overall, our work provides a novel unified and biologically-plausible paradigm for more interpretable and effective representation of ViT ANNs.
\end{abstract}

\section{Introduction}\label{introduction}
The human brain is a hugely complex, highly recurrent, robust, and remarkably efficient nonlinear neural network \cite{deneve2017brain}. Human behavior, cognition, and neural activities critically depend on the graph structure of large-scale brain connectome \cite{lynn2020humans,tompson2019individual}. Likewise, representation and predictive capability of an artificial neural network (ANN), such as convolutional neural network (CNN), 
are also closely related to the model's graph structure \cite{xie2019exploring,you2020graph}. The authors in \cite{you2020graph} hypothesized that biological neural networks (BNNs) (e.g., macaque's brain network graphs) and ANNs (CNN's relational graphs) might share common graph properties. Recently, neural networks of transformers have received extensive attention, and a variety of vision transformers (ViTs) have been developed rapidly in the computer vision field. Tolstikhin et al. \cite{tolstikhin2021mlp} and Yu et al. \cite{yu2021metaformer} summarized the message passing in transformers as Token Mixer and Channel MLP (multilayer perceptron). The core idea of various ViT models is the well-designed Token Mixer which controls the information communication between spatial tokens, while most ViTs follow similar design of the Channel MLP (two fully-connected linear with GELU function), for examples, the self-attention \cite{vaswani2017attention} in ViT \cite{dosovitskiy2020image}, the shifted window attention in Swin \cite{liu2021swin}, the token-mixing MLP in Mixer \cite{tolstikhin2021mlp,touvron2021resmlp}, the pooling in MetaFormer \cite{yu2021metaformer}, and so on. Essentially, exploring ViT's graph structure (in terms of information communication), its relevance to the representation/predictive performance, and its similarity to BNNs, is of great importance for deeper understanding of ViTs and their wider applications.

The work in this paper is inspired by the relational graph constructed on CNN models \cite{you2020graph}. You and colleagues \cite{you2020graph} introduced the pioneering concept of \emph{relational graph} which considers channels of CNN features as nodes. They systematically investigated how the relational graph structure of CNN networks affects their predictive performance. They found sweet spots in the relational graphs of CNN networks, which are similar to the real BNNs reconstructed from brain science data. 
Built upon the inspiring work in \cite{you2020graph}, in this paper, we proposed the following major innovations. (1) The relational graph in \cite{you2020graph} was constructed only on the multilayer perceptron (MLP) or CNNs. Here, we will focus on defining and constructing relational graphs on ViT models. (2) The relational graph in \cite{you2020graph} only considered the communication between the dimensions or channels of CNN, thus ignoring the spatial information aggregation. Here, we will consider the spatial patch information communication in ViT as a key factor. (3) The relational graph in \cite{you2020graph} only described the network's graph topology, thus ignoring the influence of parameter weights which can reflect the ability of capturing embedding features from images. Here, we will consider the influence of parameter weights in ViT as a key factor. Overall, our work provides a novel and unified relational graph representation of different ViT architectures for systematic understanding and assessment of model representation performance. 

Specifically, inspired by MetaFormer and MLP-Mixer \cite{yu2021metaformer,tolstikhin2021mlp}, we divided the \emph{relational graph} in ViT models into \emph{aggregation graph} and \emph{affine graph}, where the \emph{aggregation graph} describes the spatial patch information interaction, while the \emph{affine graph} reflects the information communication between channels. For example, in ViT transformer, we considered the self-attention as an aggregation network and the feedforward network (FFN) as an affine network. In this work, we explored five types of widely known ViT models, including ViT \cite{dosovitskiy2020image}, DeiT\cite{pmlr-v139-touvron21a}, Swin \cite{liu2021swin}, Mixer \cite{tolstikhin2021mlp} and MetaFormer \cite{yu2021metaformer} as examples.

In order to systematically explore the advantages of ViT models, we investigated the relationships between two sets of key factors. 
(1) ViT's relational graph structure vs its predictive performance. To relate to the model parameters rather than only model's graph structure, especially when pre-training/fine-tuning has become a popular methodology for various vision tasks, more attention was paid to the feature extraction ability of the ViT model's backbone. We evaluated the classification performances of these different models with fixed backbones and graph structures. 
(2) ViT's relational graph structure vs those of BNNs. The convolutional and pooling layers in CNNs were directly inspired by the classic notions of simple cells and complex cells in visual neuroscience \cite{lecun2015deep,hubel1962receptive}. Here, we explored the similarities of spatial information communication patterns represented by the proposed relational graphs in ViT models and those in real BNNs represented by brain graphs derived from neuroscience data.
Overall, the main contributions of our work are:
\begin{itemize}
    \item We introduced a novel, unified and biologically-plausible relational graph representation of vision transformers for the first time.
    \item We discovered the sweet spots of graph measures, including clustering coefficient and average path length, on the relational graphs of various ViT models with significantly improved predictive performances.
    \item We found that the proposed relational graph representation of ViT has high similarity with real BNNs derived from brain science data. 
\end{itemize}

\section{Related Work}

\paragraph{Vision Transformer} \label{Vision Transformer}
Vision transformer has developed rapidly in recent years from the ViT \cite{dosovitskiy2020image} to the recent SSA \cite{ren2021shunted}, Uniformer \cite{li2022uniformer}, ViT-G/14 \cite{DBLP:journals/corr/abs-2106-04560} and so on. With no or little introduction of inductive bias, ViT \cite{DBLP:journals/corr/abs-2106-04560} can achieve satisfactory performance in various computer vision tasks, despite some difficulties in training. Some studies have improved it by effective model training skills or introducing inductive bias. For example, Touvron et al. \cite{pmlr-v139-touvron21a} introduced distillation learning with distillation token to ViT for data-efficient training. Chen et al. \cite{chen2021crossvit} proposed CrossViT for multi-scale images. Liu et al. \cite{liu2021swin} designed a windows attention for greater efficiency by limiting self-attention computation to non-overlapping local windows. Li et al. \cite{li2022uniformer} combined the advantage of CNN and ViT and proposed UniFormer which aggregated local information via convolution at shallow layers and global information via self-attention at deep layers. Tolstikhin et al. \cite{tolstikhin2021mlp} introduced a pure MLP structure for computer vision and applied a token mixer MLP for information communication between the tokens. Inspired by Mixer, Yu et al. \cite{yu2021metaformer} proposed a MetaFormer which adopted an average pool kernel as the aggregation function instead of self-attention or MLP. One of the core characteristics of these ViTs is to design different attention strategies, which act as different aggregation functions for information communication between tokens. However, there is still a key lack of unified representation of different aggregation strategies in ViT models for systematic understanding and assessment of model representation/predictive performance. Inspired by the relational graph \cite{you2020graph}, we aim to propose a unified relational graph representation of ViTs.

\paragraph{Relational Graph}
The concept of relational graph in neural networks was first proposed in \cite{you2020graph}, where the message exchanges along the graph structure of neural networks. By regarding multiple dimensions or channels of CNN features as  
nodes in the graph, You et al. \cite{you2020graph} converted different random graphs into different structural neural networks in MLP or CNN models. The proposed relational graph has satisfactory generalizability and similarity with real BNNs when compared with the computational graph \cite{xie2019exploring}. More interestingly, the authors in \cite{you2020graph} found that the highest performing model tends to have a similar graph measure which is called the sweet spot. However, the proposed relational graph still has certain limitations as discussed in Section~\ref{introduction}, and we aim to introduce a more general and biologically-plausible relational graph representation for ViTs.

\paragraph{BNNs and Graphs}
Modeling and interpreting the brain as a complex networked system provides remarkable novel insights into a series of fundamental neuroscience questions, e.g., how the brain organizes its structure and function; how spatially distributed brain regions interact during cognitive processes; and how the structural and functional organizations alter in brains with neurological, psychiatric, or psychological diseases\cite{stam2009graph}. Among these studies, graph theoretical analysis of a neural network describes the topological attributes regarding to its adaptability, robustness, general cost and efficiency and etc. A neural network is typically represented as a graph $G=(V, E)$, where $V$ is the set of nodes representing brain regions or neurons, and $E$ is the set of edges revealing the functional (e.g., synchronization of functional activities), structural (e.g., neural fibers or synapses) or effective (causal relationship) connectivities between node-pairs. With such a configuration, graph measures of neural networks can be quantified by borrowing concepts and tools in the field of graph theory. Although the real BNNs derived from brain science data have profoundly inspired the ANNs, how these well-performing ANNs such as ViTs are similar to those of BNNs are largely unexplored and unknown.

\section{Relational Graph in Vision Transformers}

\subsection{Structure of Vision Transformer Models}
\begin{figure}
\centering
\includegraphics[scale=0.4]{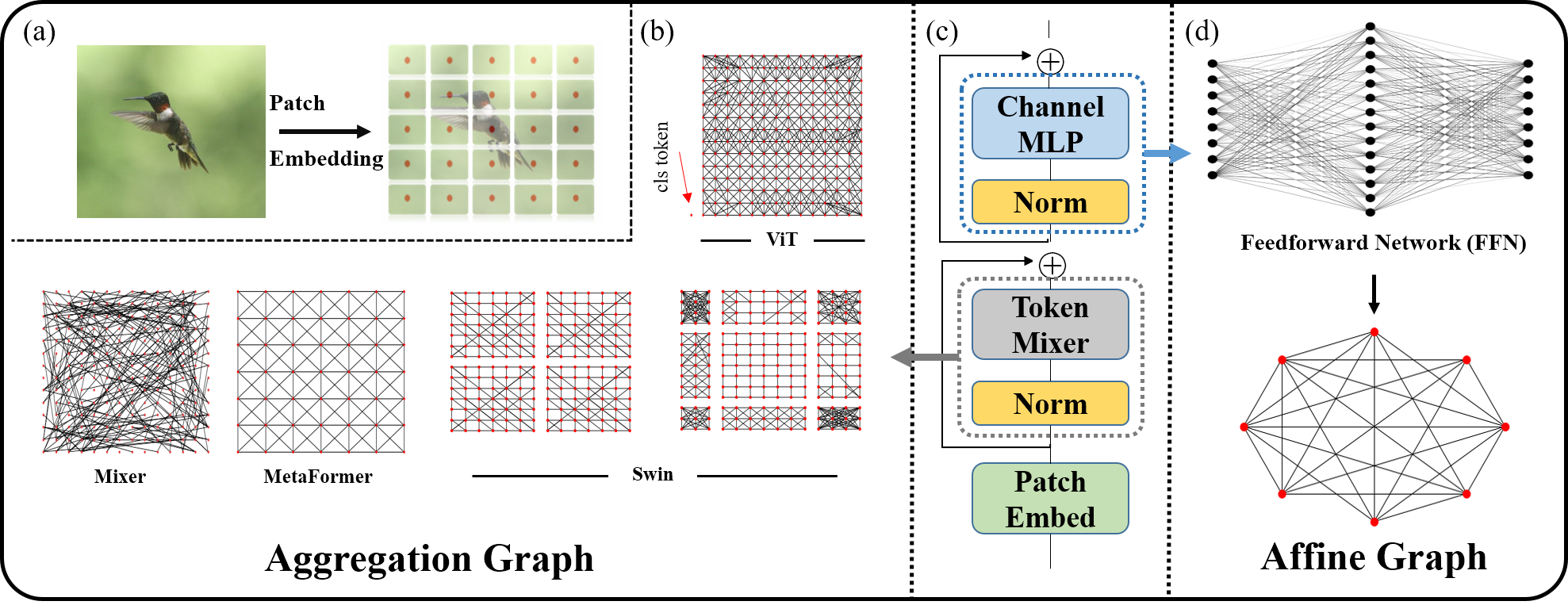}
\caption{The aggregation and affine graph in different ViT models. (a) The patch embedding in vision transformer models. (b) The aggregation graph and (d) the affine graph in vision transformer models. (c) the structure of transformer vision models.}
\label{fig1}
\end{figure}

As shown in Figure \ref{fig1}(c), there are three core parts in vision transformers: Patch Embedding~\ref{fig1}(a), Token Mixer~\ref{fig1}(b) and Channels MLP~\ref{fig1}(d). The Patch Embedding layer projects the input images into non-overlapping image tokens as word tokens in NLP. The Token Mixer allows information communication between different spatial locations (images tokens) while the Channels MLP allows information communication between different channels. Therefore, the overall structure of vision transformers is represented as:
\begin{align}
    &X = PatchEmbed(Image)  \\
    &Y = TokenMixer(Norm(X))+X  \\
    &Z = ChnnelsMLP(Norm(Y))+Y
\end{align}
where $Norm$ is the normalization function. Note that there are also some other model designs such as the learnable absolute position embedding and class token in ViT, the relative position bias in Swin, the distillation token in DeiT, etc. Here, we only discuss the backbone of the model to simplify the issue.

\subsection{Aggregation and Affine Graphs}
Inspired by the MetaFormer model \cite{yu2021metaformer}, the two core message exchange functions of vision transformer are Token Mixer and Channel MLP. The Token Mixer aggregates information from tokens while the Channel MLP allows information communication between different channels \cite{tolstikhin2021mlp}. Therefore, we defined the relational graph of vision transformer as two key sub-graphs: Token Mixer Graph and Channel Mixer Graph, and formally annotated them as Aggregation Graph and Affine Graph according to their major roles.

\paragraph{Aggregation Graph} By considering each token as a node in the graph, the token mixer (e.g. self-attention) plays the same role as the aggregation function in graph neural networks via updating their embedding features by aggregation of the information from its neighborhoods. With this message-passing paradigm, the ViT model can globally integrate information from different regions (tokens) in each layer, whereas for CNN models, message passing is restricted to the kernel size. For example, the propagation function between tokens of MLP-Mixer is written as:
\begin{equation} \label{Token Mixer}
    x = x + (LN(x^T)W)^T
\end{equation}
where $LN$ is the LayerNorm function and $W$ is the set of learning parameters. If we regard each token as a node in graph, this function is written as
$TokenMixer(x) = W^T x I$.
Therefore, the normalized adjacent matrix of aggregation graph of Mixer is $Softmax(W^T/\sqrt{dim})$. Note that we used the same normalization function as \cite{dosovitskiy2020image,vaswani2017attention}, which can also be written as $\hat{A}=AD^{-1}$ where $A=exp(W^T/\sqrt{dim})$ and $D_{ii}=\sum_j A_{ij})$.

\paragraph{Affine Graph} By considering each channel as a node in the graph, the channel mixer allows the message passing along channels in the affine graph. Different from the aggregation graph which conveys spatial information of tokens, the affine graph is designed for the information communication between different channels. The channel mixer always contains two full-connection layers and a nonlinear activation function as:
\begin{equation}
    x = x +\delta(LN(x)W_1)W_2
\end{equation}
where $\delta$ is the nonlinear activation function such as GELU and ReLU. Therefore, the adjacent matrix of affine graph is written as $Softmax({W_1W_2}/\sqrt{dim})$.

\subsection{Construction of Aggregation Relational Graph in Vision Transformers}

Since the key difference among different ViT models is the different design of Token Mixer, i.e., aggregation graph, we therefore mainly focused on a unified aggregation relational graph representation of different ViT models, instead of an affine one in this study. Specifically, we selected ViT\cite{dosovitskiy2020image}, DeiT\cite{pmlr-v139-touvron21a}, Swin\cite{liu2021swin}, Mixer\cite{tolstikhin2021mlp} and MetaFormer\cite{yu2021metaformer} as five representative examples among all ViT models as introduced in Related Work \ref{Vision Transformer} due to the space limit.

\paragraph{ViT/DeiT} The ViT \cite{dosovitskiy2020image} and DeiT \cite{pmlr-v139-touvron21a} models both follow a pure transformer \cite{vaswani2017attention} design while more training schemes are adopted in DeiT. Both ViT and DeiT adopt the multi-head self-attention (MHSA) as the Token Mixer, which can be described as:
\begin{equation}
    Softmax(xW_qW_k^Tx^T/\sqrt{dim})xW_v
\end{equation}
In addition, they keep the position information with learnable absolute position embedding $x=x+P$ where $P$ is the position embedding. For the sake of analysis, we assumed that different tokens are independent of each other, and the information communication between tokens is carried out through position embedding. Therefore, the adjacent matrix of aggregation graph is written as:
\begin{equation} \label{ViT Aggre}
    \hat{A} = Softmax({PW_q W_k^T P^T}/\sqrt{dim})
\end{equation}
It is worth noting that we did not use the information of the $W_v$ matrix due to its role as a feature projection rather than an exchange of spatial information between tokens.

\paragraph{Swin} Compared with ViT, Swin \cite{liu2021swin} proposed the windows attention and shifted windows attention in order to introduce inductive bias into vision transformer. Additionally, relative position bias rather than absolute position embedding is applied in Swin. The Token Mixer in windows of Swin is:
\begin{equation}
    Softmax(xW_qW_k^Tx^T/\sqrt{dim}+B+Mask)xW_v
\end{equation}
where $B$ is the relative position bias and $Mask$ is designed for shifted attention operation, which is window-specific. Therefore, the aggregation graph in each window is:
\begin{equation}
    \hat{A} = Softmax(I/\sqrt{dim}+B+Mask)
\end{equation}
where $I$ is the unit matrix.

\paragraph{Mixer} The MLP-Mixer is an MLP only vision model \cite{tolstikhin2021mlp,touvron2021resmlp}, primarily composed of token-mixing MLP and channel-mixing MLP. The token-mixing MLP acts on the transpose matrix of the image token, thus aggregating spatial information between different tokens while the channel-mixing MLP is the same as the FFN in transformer. Note that there is no position embedding in Mixer as the token-mixing MLPs are sensitive to the order of the input tokens. Based on the token mixer defined in Eq.\ref{Token Mixer}, the aggregation graph is written as:
\begin{equation}
    \hat{A} = Softmax(W/\sqrt{dim})
\end{equation}
where $W$ is the set of learning parameters of MixingToken in Eq.\ref{Token Mixer}.

\paragraph{MetaFormer} Based on the MLP-Mixer, Yu et al. \cite{yu2021metaformer} proposed a PoolFormer which uses pooling functions to perform information communication between tokens and greatly reduces the computational effort and parameters of the model. Therefore, the aggregation graph of MetaFormer is:
\begin{equation} \label{pool}
    a_{i,j} = 1/K^2,j\in Ner(i)
\end{equation}
where $K$ is the kernel size and $Ner(*)$ is the neighborhood set.

\subsection{Aggregation Graph between Layers}

The considerable difference in depth and width of different ViT models makes it challenging to adopt a unified framework for model assessment. You \cite{you2020graph} assumed multiple features as one node in order and layers as rounds, thus making different models comparable. Inspired by this study, we fixed the aggregation graph size as $14 \times 14$(+1) nodes and also regarded the layers as rounds. Therefore, for the high resolution aggregation graph, we down-sampled it by:
\begin{equation}
    Downsample(\Hat{A}_{xy}) = 1/K * \sum_i^{Ner(x)} \sum_j^{Ner(y)} \Hat{A}_{ij} 
\end{equation}
where $K$ is the down-sampling rate and $Ner(*)$ is the neighborhood set in the high resolution image. For the low resolution aggregation graph, we up-sampled it by:
\begin{equation}
    Upsample(\Hat{A}_{xy}) = 1/K \Hat{A}_{ij},i=x//K,j=y//K
\end{equation}
where $i,j$ is the index of the raw resolution image and $x,y$ is the index of downsampled or upsampled image. The final aggregation graph of a model is:
\begin{equation}
    Final \hat{A} = \prod^{Layers} Sampled (\hat{A})
\end{equation}
Note that we normalized ($Softmax(Final \hat{A})$) matrix for ease of analysis.

\section{Experiments}
Following the suggestion in \cite{you2020graph}, we adopted two graph measures, the average path length and clustering coefficient that characterize the integration and segregation of a network respectively, to systematically explore (1) the relationship between these graph measures and downstream task performances of ViT models, (2) the effectiveness of these graph measures as ViT model training indicators from scratch, and (3) the topological similarity between ViT models and real BNNs.

\subsection{Downstream Task of Pretrained ViT Models}

\paragraph{Datasets}
We used image classification as the downstream task and evaluated the performance of the model's backbone on three different and widely-known datasets: CIFAR10 \cite{krizhevsky2009learning} including 50K training images and 10K validation images, Animal10 \cite{song2019selfie} including 50k training images and 5k validation images, and Flower17 \cite{Nilsback06} including 80 images (60 for training, 10 for testing and 10 for validation) for each category.

\paragraph{Models}
We adopted 21 publicly available pretrained models of the five ViTs (ViT, DeiT, Swin, Mixer, and MetaFormer) provided in timm\footnote{https://github.com/rwightman/pytorch-image-models}. The details of all 21 pretrained ViT models are provided in the Appendix A.

\paragraph{Settings}\label{setting}
We followed the same setting of \cite{ridnik2021imagenet} for better model training\footnote{https://github.com/Alibaba-MIIL/ImageNet21K}. An SGD optimizer with a weight decay of 0.0001 was applied. We employed an OneCycleLR \cite{smith2019super} learning rate scheduler and the initial(max) learning rate was 0.0005(0.001) with cosine annealing. The percentage of the cycle spent increasing the learning rate was set to 0.1. The label smoothing of CrossEntropy Loss was set to 0.2 and batch size was 128/256.
The data augmentations included Cutout \cite{devries2017improved} and RandAugment \cite{cubuk2020randaugment}. All images were resized to $224 \times 224$ in consistent with the pretrained model. All models were with pretrained parameter weights and only the parameters of classification header would change while the backbone was fixed.

\subsection{ViT Model Training from Scratch}
\label{train from scratch}
To explore the dynamic changing characteristics of the two graph measures during model training and their effectiveness as ViT model training indicators, we trained the ViT-Ti \cite{dosovitskiy2020image} from scratch on ImageNet-1k \cite{ILSVRC15} using the same setting as in \ref{setting} and recorded the accuracy and graph measures for each epoch. We used Adam optimizer with batch size of 800 for better performance.

\subsection{Graph Similarities between ViTs and BNNs}
A variety of real BNNs were compared to the ViT relational graphs, including the synaptic connectomes of the anterior (Wrom\_279) and posterior (Worm\_269) nervous system of the C. elegans; the synaptic connectomes for the pharyngeal nervous systems of two nematodes with divergent feeding behavior (Worm\_54 and Worm\_50); the neuronal connectomes of the rat brain revealed by neuronal pathway tracers (Rat\_493, Rat\_503a and Rat\_503b); the macroscopic brain network of cat as reconstructed from tract tracing data (Cat\_65); the structural connectome of a macaque monkey derived from a collation of tract tracing studies (Macaque\_242); the interareal connectivities of macaque monkeys (Macaque\_V91, Macaque\_CC91 and Macaque\_93) revealed by retrograde tracers. A detailed description of those BNNs is referred to \footnote{https://neurodata.io/project/connectomes/}. The Euclidean distance between the graph measures of BNNs and ViTs was calculated as the topological similarity between BNNs and ViTs.

\section{Result}

\subsection{Relationship between Aggregation Graph Measures and Classification Performances}

\begin{figure}[h]
    \centering
    \includegraphics[scale=0.36]{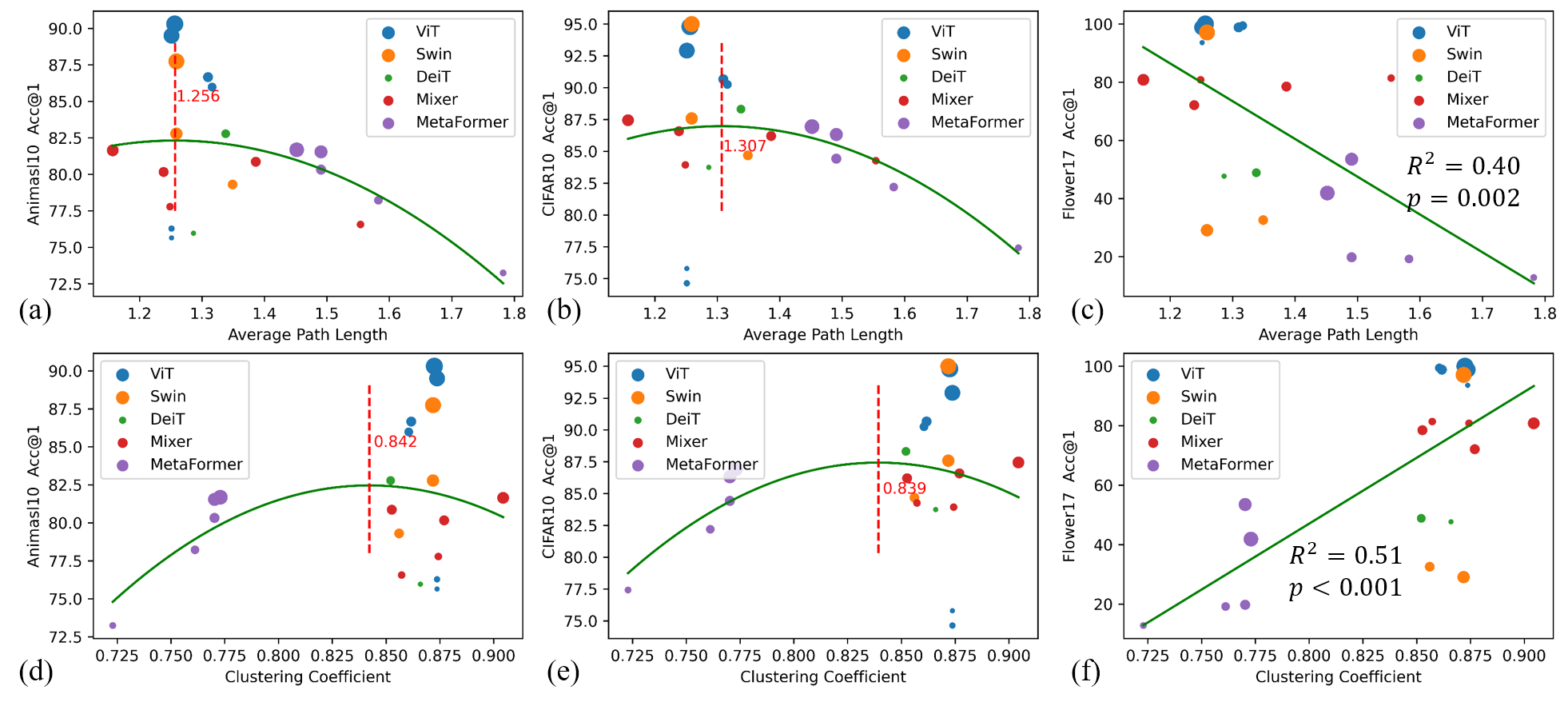}
    \caption{Relationship between two aggregation graph measures and classification performance of 21 ViT models on three datasets. Larger spot size represents larger model parameters.}
    \label{aggregation_dim}
\end{figure}

\begin{wrapfigure}{r}{7cm}
\includegraphics[width=7cm]{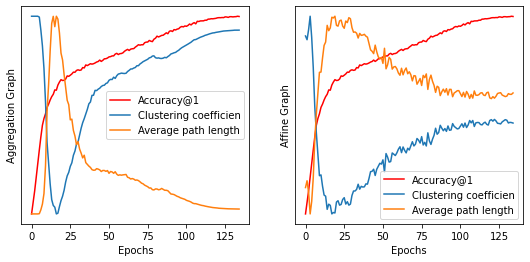}
\captionof{figure}{Dynamic change of the two graph measures during ViT-Ti model training from scratch on ImageNet-1k dataset.}\label{graph changing}
\end{wrapfigure}
\paragraph{Identification of Sweet Spot} There was a smooth U-shape correlation between the model classification performance and each of the two graph measures in the first two datasets using a second degree polynomial regression as illustrated in Figure ~\ref{aggregation_dim}(a,b,d,e). Similarly in \cite{you2020graph}, we successfully identified stable sweet spots $C \in [0.839,0.842]$ and $ L \in [1.256,1.307]$ across the two different datasets as highlighted as red dashed lines in Figure \ref{aggregation_dim}(a,b,d,e) corresponding to the best classification performance.

\paragraph{Indicator of Model Adaptability} As illustrated in Figure \ref{aggregation_dim}(c,f), the relationship between the graph measure and classification performance in the third dataset (Flower17) exhibited a significant linear correlation instead of a U-shape one. The interpretation was that the Flower17 dataset has only 60 images per category similar to few-shot learning task. Therefore, these two graph measures of the proposed relational graph not only helped us identify the sweet spot with large sample data, but also reflected the adaptive learning ability of the model with few sample data. In summary, our results illustrated that the two graph measures can be used as effective indicators of model classification performance.


\subsection{Dynamic Changing Characteristics of Graph Measures During Model Training}
\label{SecGraphChange}
\paragraph{Indicator of Model Learning} We averaged the graph measures of each layer as the representation of the final graph. A threshold of 1/192 (embedding dim/node numbers) for the affine graph and 1/197 (token numbers) for the aggregation graph was selected to illustrate the dynamic change of graph structure during model training. Note that we concatenated all aggregation graphs along the diagonal rather than through rounds in order to better highlight the changing trend of the model. We reported the top-1 accuracy on validation dataset and two graph measures in Figure. \ref{graph changing}. Interestingly, except for the initial 25 epochs, there was considerable consistency between the graph measure and the accuracy of the model, indicating that the two graph measures may also serve as a more general and effective indicator of model training.

\subsection{Graph Measure Similarity between ViTs and BNNs}
\label{SimilaritywithBNN}
\begin{figure}[ht]
    \centering
    \includegraphics[scale=0.4]{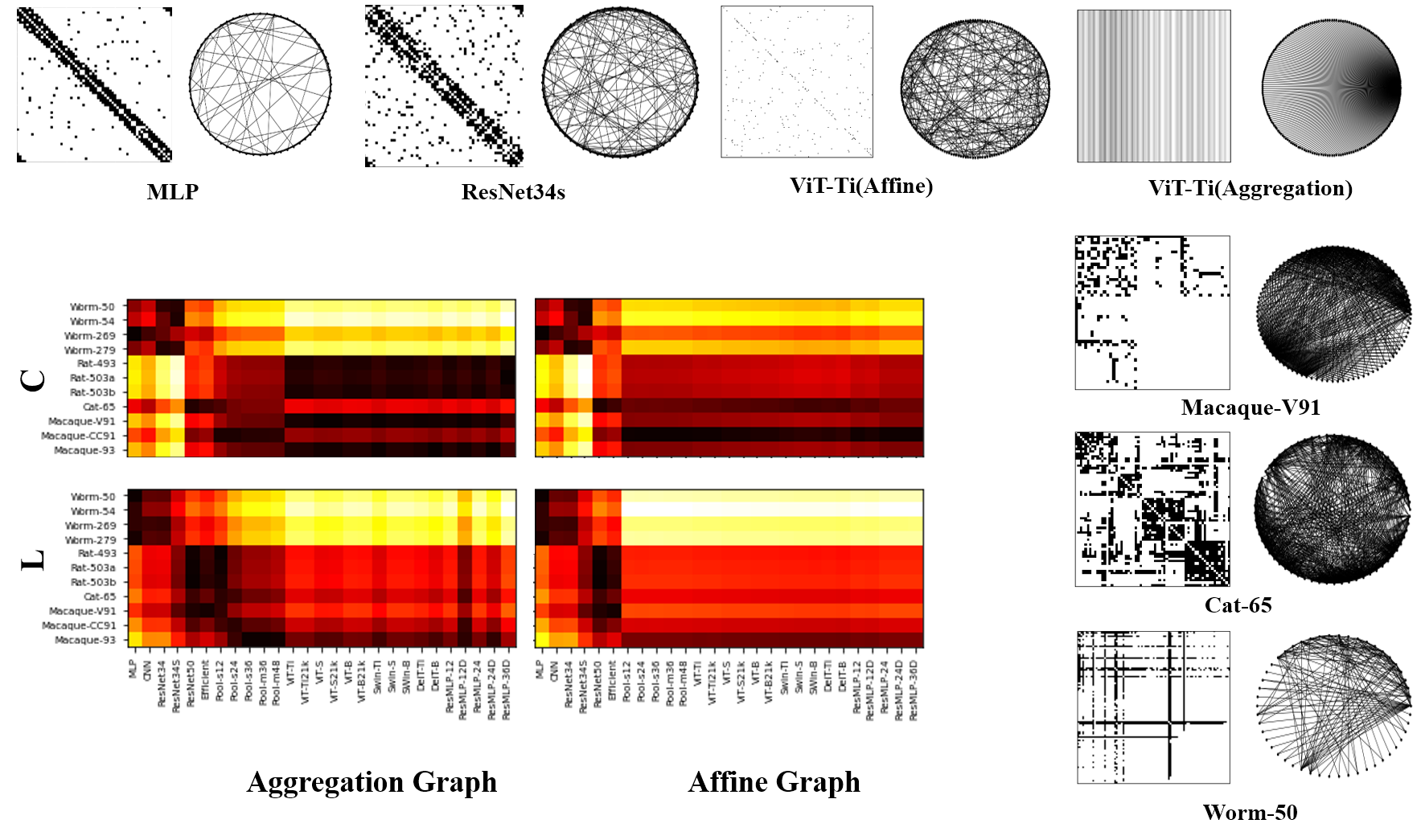}
    \caption{Similarity of relational graphs between vision models and BNNs.}
    \label{fig-brain}
\end{figure}

The graph measure similarity between ViT models and BNNs is shown in Figure \ref{fig-brain}. Note that the relational graphs of MLP and CNN based models are from \cite{you2020graph}. We see that the graph measures of ViTs were close to the rat, cat and macaque's brain neural networks, while those of the MLP and CNN based models are similar to worms. This inspiring result might suggest the superiority of ViT over MLP/CNN in terms of information communication and exchange efficiency, given that mammalian brains such as the rat, cat and macaque are considered to be much more advanced and optimized than the worm's neural networks.

\section{Discussion}

\subsection{Fine-tune Partial vs. Full Model}
\begin{figure}[ht]
\centering
\begin{minipage}[t]{0.48\linewidth}
\centering
\includegraphics[scale=0.3]{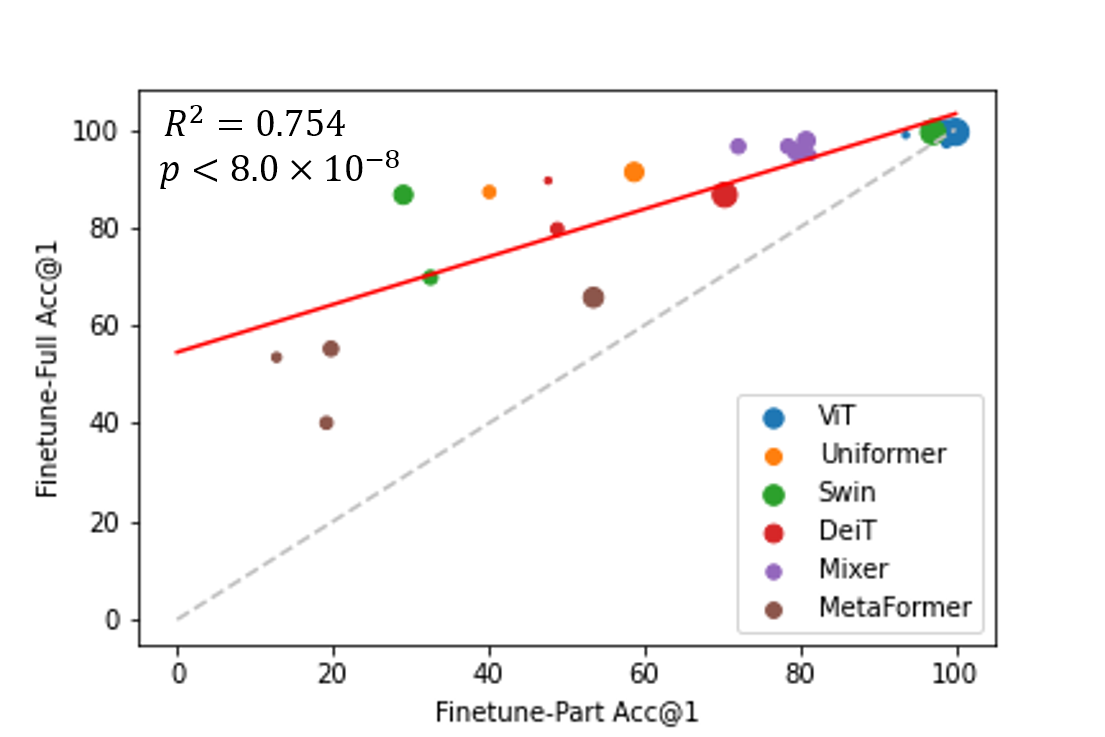}
\captionof{figure}{The prediction accuracy on Flower17 with fine-tuning partial and full models.}
\label{fig-finefullvspart}
\end{minipage}
\hfill
\begin{minipage}[t]{0.48\linewidth}
\centering
\includegraphics[scale=0.5]{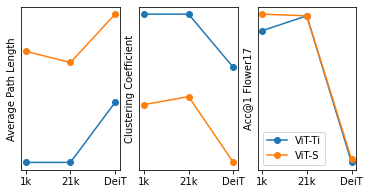}
\captionof{figure}{The graph measures of different training methods in the same model.}
\label{fig-Training }
\end{minipage}
\end{figure}

The construction of the proposed relational graph was affected by ViT model parameters. To keep consistent with the graph structure, we froze the parameters of ViT model's backbone and only updated the classification header. To justify weather fixed backbone would affect the assessment of model performance, we provided the predictive performance on Flower17 dataset with/without freezing the backbone parameters in Figure \ref{fig-finefullvspart}. We observed that the performance with/without freezing has a high linear correlation, indicating that it is reasonable to assess the model performance by graph measures with fixed backbone.

\subsection{Graph Structures with Different Sampling Efficiencies}

Section \ref{SecGraphChange} demonstrated that the graph measures changed during model training with the same training setting. We further reported the graph measure changes of ViT with different training strategies such as training with more datasets or with distillation learning in Figure~6. Compared with DeiT with more training schemes \cite{pmlr-v139-touvron21a}, the graph measures of models pretrained with more data (ImageNet-21k pretrained) had higher similarity with the ImageNet-1k pretrained models. More interestingly, these graph measure changes also corresponded to the model generalizability, which decreased as the average path length increased and the clustering coefficient decreased. These findings were consistent with those in Figure~\ref{aggregation_dim} and further indicated that the graph measures could reflect not only the model classification performance, but also the model sampling efficiency.

\subsection{Aggregation Graphs between Layers}
\begin{wrapfigure}{r}{7.5cm}
\includegraphics[width=7.5cm]{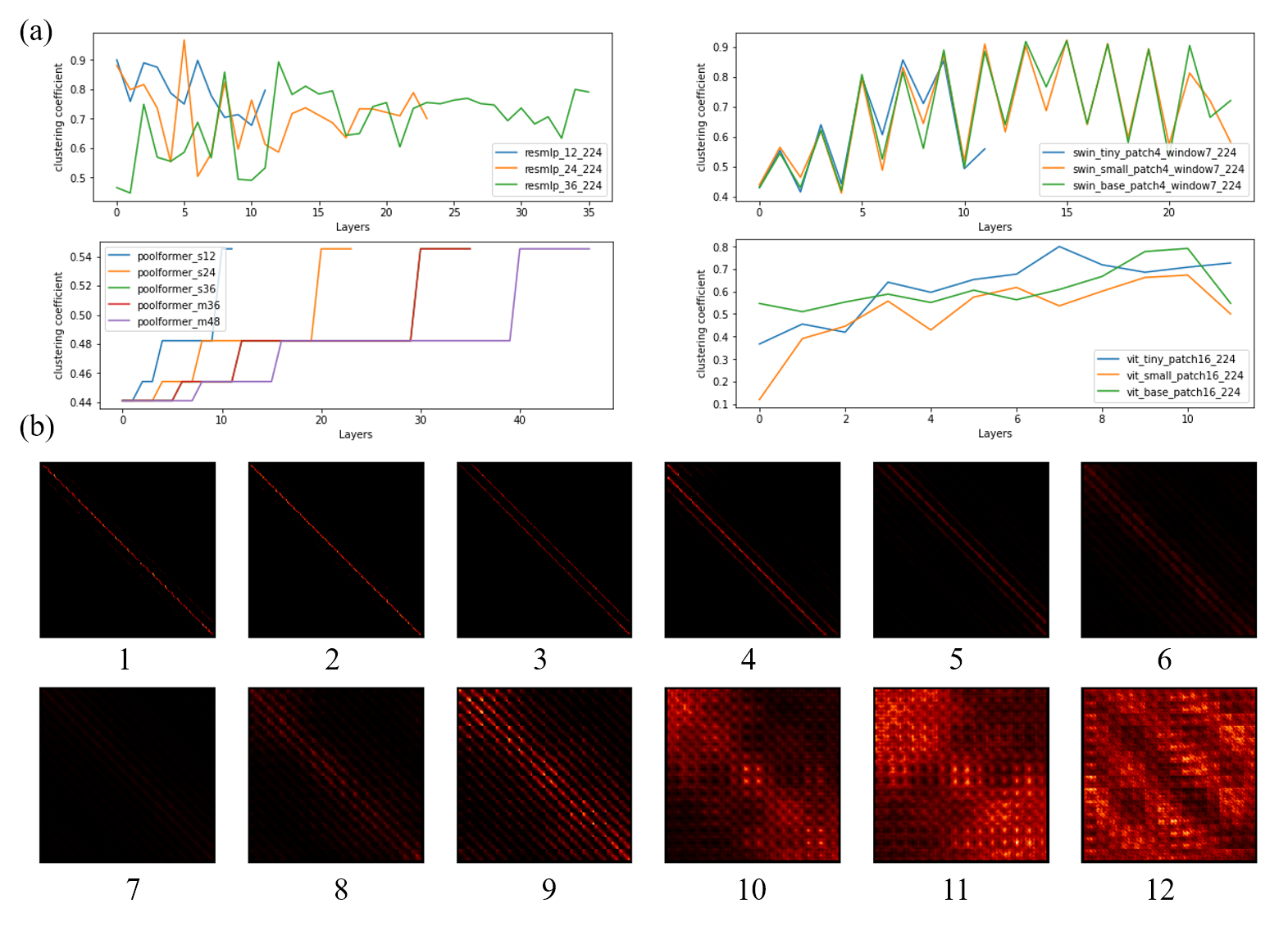}
\captionof{figure}{The aggregation graphs between layers in different ViT models. (a) The clustering coefficients of Mixer, MetaFormer, ViT and Swin. (b) The visualization of aggregation graphs in ViT-B.}\label{fig-aggregation}
\end{wrapfigure}
In addition to the overall aggregation graph of different ViT models, we also investigated the aggregation graphs of different layers of the model. As shown in Figure \ref{fig-aggregation}, the clustering coefficient tends to increase with the number of layers (Figure \ref{fig-aggregation}(a)), which was further demonstrated by a visible example (ViT-B) in Figure \ref{fig-aggregation}(b). The clustering coefficients reflected the degree of aggregation of the network that low-level features continuously aggregated into high-level features as the model layers became deeper. Our findings are consistent with previous studies \cite{raghu2021vision} from a graph measure perspective.

\subsection{Impact on Model Design and Optimization of ANNs}
The proposed relational graph provides a unified paradigm to represent various ViT models, thus enabling effective comparisons among different ANNs and their associations with BNNs. Given that the brains are already highly optimized BNNs and these structural and functional BNNs possess a variety of interesting and nice graph properties, such as the aforementioned sweet spots, we premise that BNNs' graph characteristics could provide potential guidance for the design and optimization of ANNs, e.g., through neural architecture search (NAS), and could offer benchmarks for evaluating those optimized ANNs. Also, additional graph models, abstractions, and common frameworks could be explored and summarized from BNNs and then used to inform and guide the design and optimization of next-generation ANNs and their wide applications in the future.

\subsection{Limitation}
The spatial information exchange in ViT is often associated with the content of images. However, it is difficult to integrate specific data information into the graph construction of networks. Here, we made an assumption that the patches of images are independent of each other and the spatial relationship is retained only through positional information. However, some models such as Mixer \cite{tolstikhin2021mlp} and ResMLP \cite{touvron2021resmlp} did not use position embedding and might be sensitive to the order of the input tokens \cite{tolstikhin2021mlp}. Therefore, integrating image content into the model structure deserves future exploration.


\section{Conclusion}
In this paper, we proposed a novel unified and biologically-plausible relational graph representation of vision transformers. By decomposing the information transfer graph of the network into the aggregation and affine graphs, our method can be applied to almost all representative visual models. By exploring the differences in graph measures of aggregation graphs in different ViTs, we found that the model performance is closely related to the graph measures, especially when the sample size is small. We also found that the proposed relational graph representation of ViTs has high similarity with real BNNs derived from brain science data. Overall, we provided an interpretable and effective way for analyzing ViT models and linking models' relational graphs with BNNs, and offered novel insights on the design of ANNs.


\begin{thebibliography}{10}

\bibitem{deneve2017brain}
Sophie Den{\`e}ve, Alireza Alemi, and Ralph Bourdoukan.
\newblock The brain as an efficient and robust adaptive learner.
\newblock {\em Neuron}, 94(5):969--977, 2017.

\bibitem{lynn2020humans}
Christopher~W Lynn and Danielle~S Bassett.
\newblock How humans learn and represent networks.
\newblock {\em Proceedings of the National Academy of Sciences},
  117(47):29407--29415, 2020.

\bibitem{tompson2019individual}
Steven~H Tompson, Ari~E Kahn, Emily~B Falk, Jean~M Vettel, and Danielle~S
  Bassett.
\newblock Individual differences in learning social and nonsocial network
  structures.
\newblock {\em Journal of Experimental Psychology: Learning, Memory, and
  Cognition}, 45(2):253, 2019.

\bibitem{xie2019exploring}
Saining Xie, Alexander Kirillov, Ross Girshick, and Kaiming He.
\newblock Exploring randomly wired neural networks for image recognition.
\newblock In {\em Proceedings of the IEEE/CVF International Conference on
  Computer Vision}, pages 1284--1293, 2019.

\bibitem{you2020graph}
Jiaxuan You, Jure Leskovec, Kaiming He, and Saining Xie.
\newblock Graph structure of neural networks.
\newblock In {\em International Conference on Machine Learning}, pages
  10881--10891. PMLR, 2020.

\bibitem{tolstikhin2021mlp}
Ilya~O Tolstikhin, Neil Houlsby, Alexander Kolesnikov, Lucas Beyer, Xiaohua
  Zhai, Thomas Unterthiner, Jessica Yung, Andreas Steiner, Daniel Keysers,
  Jakob Uszkoreit, et~al.
\newblock Mlp-mixer: An all-mlp architecture for vision.
\newblock {\em Advances in Neural Information Processing Systems}, 34, 2021.

\bibitem{yu2021metaformer}
Weihao Yu, Mi~Luo, Pan Zhou, Chenyang Si, Yichen Zhou, Xinchao Wang, Jiashi
  Feng, and Shuicheng Yan.
\newblock Metaformer is actually what you need for vision.
\newblock {\em arXiv preprint arXiv:2111.11418}, 2021.

\bibitem{vaswani2017attention}
Ashish Vaswani, Noam Shazeer, Niki Parmar, Jakob Uszkoreit, Llion Jones,
  Aidan~N Gomez, {\L}ukasz Kaiser, and Illia Polosukhin.
\newblock Attention is all you need.
\newblock {\em Advances in neural information processing systems}, 30, 2017.

\bibitem{dosovitskiy2020image}
Alexey Dosovitskiy, Lucas Beyer, Alexander Kolesnikov, Dirk Weissenborn,
  Xiaohua Zhai, Thomas Unterthiner, Mostafa Dehghani, Matthias Minderer, Georg
  Heigold, Sylvain Gelly, et~al.
\newblock An image is worth 16x16 words: Transformers for image recognition at
  scale.
\newblock {\em arXiv preprint arXiv:2010.11929}, 2020.

\bibitem{liu2021swin}
Ze~Liu, Yutong Lin, Yue Cao, Han Hu, Yixuan Wei, Zheng Zhang, Stephen Lin, and
  Baining Guo.
\newblock Swin transformer: Hierarchical vision transformer using shifted
  windows.
\newblock In {\em Proceedings of the IEEE/CVF International Conference on
  Computer Vision}, pages 10012--10022, 2021.

\bibitem{touvron2021resmlp}
Hugo Touvron, Piotr Bojanowski, Mathilde Caron, Matthieu Cord, Alaaeldin
  El-Nouby, Edouard Grave, Gautier Izacard, Armand Joulin, Gabriel Synnaeve,
  Jakob Verbeek, et~al.
\newblock Resmlp: Feedforward networks for image classification with
  data-efficient training.
\newblock {\em arXiv preprint arXiv:2105.03404}, 2021.

\bibitem{pmlr-v139-touvron21a}
Hugo Touvron, Matthieu Cord, Matthijs Douze, Francisco Massa, Alexandre
  Sablayrolles, and Herve Jegou.
\newblock Training data-efficient image transformers \&amp; distillation through
  attention.
\newblock In Marina Meila and Tong Zhang, editors, {\em Proceedings of the 38th
  International Conference on Machine Learning}, volume 139 of {\em Proceedings
  of Machine Learning Research}, pages 10347--10357. PMLR, 18--24 Jul 2021.

\bibitem{lecun2015deep}
Yann LeCun, Yoshua Bengio, and Geoffrey Hinton.
\newblock Deep learning.
\newblock {\em nature}, 521(7553):436--444, 2015.

\bibitem{hubel1962receptive}
David~H Hubel and Torsten~N Wiesel.
\newblock Receptive fields, binocular interaction and functional architecture
  in the cat's visual cortex.
\newblock {\em The Journal of physiology}, 160(1):106, 1962.

\bibitem{ren2021shunted}
Sucheng Ren, Daquan Zhou, Shengfeng He, Jiashi Feng, and Xinchao Wang.
\newblock Shunted self-attention via multi-scale token aggregation.
\newblock {\em arXiv preprint arXiv:2111.15193}, 2021.

\bibitem{li2022uniformer}
Kunchang Li, Yali Wang, Junhao Zhang, Peng Gao, Guanglu Song, Yu~Liu, Hongsheng
  Li, and Yu~Qiao.
\newblock Uniformer: Unifying convolution and self-attention for visual
  recognition.
\newblock {\em arXiv preprint arXiv:2201.09450}, 2022.

\bibitem{DBLP:journals/corr/abs-2106-04560}
Xiaohua Zhai, Alexander Kolesnikov, Neil Houlsby, and Lucas Beyer.
\newblock Scaling vision transformers.
\newblock {\em CoRR}, abs/2106.04560, 2021.

\bibitem{chen2021crossvit}
Chun-Fu~Richard Chen, Quanfu Fan, and Rameswar Panda.
\newblock Crossvit: Cross-attention multi-scale vision transformer for image
  classification.
\newblock In {\em Proceedings of the IEEE/CVF International Conference on
  Computer Vision}, pages 357--366, 2021.

\bibitem{stam2009graph}
CJ~Stam, W~De~Haan, ABFJ Daffertshofer, BF~Jones, I~Manshanden, Anne-Marie van
  Cappellen~van Walsum, Teresa Montez, JPA Verbunt, JC~De~Munck, BW~Van~Dijk,
  et~al.
\newblock Graph theoretical analysis of magnetoencephalographic functional
  connectivity in alzheimer's disease.
\newblock {\em Brain}, 132(1):213--224, 2009.

\bibitem{krizhevsky2009learning}
Alex Krizhevsky, Geoffrey Hinton, et~al.
\newblock Learning multiple layers of features from tiny images.
\newblock 2009.

\bibitem{song2019selfie}
Hwanjun Song, Minseok Kim, and Jae-Gil Lee.
\newblock {SELFIE}: Refurbishing unclean samples for robust deep learning.
\newblock In {\em ICML}, 2019.

\bibitem{Nilsback06}
Maria-Elena Nilsback and Andrew Zisserman.
\newblock A visual vocabulary for flower classification.
\newblock In {\em IEEE Conference on Computer Vision and Pattern Recognition},
  volume~2, pages 1447--1454, 2006.

\bibitem{ridnik2021imagenet}
Tal Ridnik, Emanuel Ben-Baruch, Asaf Noy, and Lihi Zelnik-Manor.
\newblock Imagenet-21k pretraining for the masses.
\newblock {\em arXiv preprint arXiv:2104.10972}, 2021.

\bibitem{smith2019super}
Leslie~N Smith and Nicholay Topin.
\newblock Super-convergence: Very fast training of neural networks using large
  learning rates.
\newblock In {\em Artificial intelligence and machine learning for multi-domain
  operations applications}, volume 11006, page 1100612. International Society
  for Optics and Photonics, 2019.

\bibitem{devries2017improved}
Terrance DeVries and Graham~W Taylor.
\newblock Improved regularization of convolutional neural networks with cutout.
\newblock {\em arXiv preprint arXiv:1708.04552}, 2017.

\bibitem{cubuk2020randaugment}
Ekin~D Cubuk, Barret Zoph, Jonathon Shlens, and Quoc~V Le.
\newblock Randaugment: Practical automated data augmentation with a reduced
  search space.
\newblock In {\em Proceedings of the IEEE/CVF Conference on Computer Vision and
  Pattern Recognition Workshops}, pages 702--703, 2020.

\bibitem{ILSVRC15}
Olga Russakovsky, Jia Deng, Hao Su, Jonathan Krause, Sanjeev Satheesh, Sean Ma,
  Zhiheng Huang, Andrej Karpathy, Aditya Khosla, Michael Bernstein,
  Alexander~C. Berg, and Li~Fei-Fei.
\newblock {ImageNet Large Scale Visual Recognition Challenge}.
\newblock {\em International Journal of Computer Vision (IJCV)},
  115(3):211--252, 2015.

\bibitem{raghu2021vision}
Maithra Raghu, Thomas Unterthiner, Simon Kornblith, Chiyuan Zhang, and Alexey
  Dosovitskiy.
\newblock {Do vision transformers see like convolutional neural networks?}.
\newblock {\em Advances in Neural Information Processing Systems}, 34, 2021.

\end{thebibliography}



\end{document}